\documentclass[letterpaper, 10 pt, conference]{ieeeconf}  % Comment this line out if you need a4paper
\pdfoutput=1 

\IEEEoverridecommandlockouts                              % This command is only needed if 
\usepackage{graphics} % for pdf, bitmapped graphics files
\usepackage{graphicx}
\usepackage{wrapfig}
\usepackage[font=footnotesize]{caption}
\usepackage{subcaption}
\usepackage{color}
\usepackage[T1]{fontenc}
\usepackage{dblfloatfix}

\usepackage{mathptmx} % assumes new font selection scheme installed
\usepackage{times} % assumes new font selection scheme installed
\usepackage{amsmath} % assumes amsmath package installed
\usepackage{amssymb}  % assumes amsmath package installed
\usepackage{mathtools}
\newcommand{\norm}[1]{\left\lVert#1\right\rVert}

\title{\LARGE \bf
Hierarchical Bayesian Data Fusion for Robotic Platform Navigation}

\author{Andr\'{e}s F. Echeverri, Henry Medeiros, Ryan Walsh, Yevgeniy Reznichenko and Richard Povinelli % <-this % stops a space
\thanks{The authors are with the Department of Electrical and Computer Engineering, Marquette University, Milwaukee, WI, USA
        {\tt\small \{andres.echeverriguevara\}, \{henry.medeiros\}, \{ryan.w.walsh\}, \{yevgeniy.reznichenko\}. \{richard.povinelli\} @marquette.edu}}%
}

\begin{document}

\maketitle
\thispagestyle{empty}
\pagestyle{empty}

\begin{abstract}

Data fusion has become an active research topic in recent years. Growing computational performance has allowed the use of redundant sensors to measure a single phenomenon. While Bayesian fusion approaches are common in general applications, the computer vision field has largely relegated this approach. Most object following algorithms have gone towards pure machine learning fusion techniques that tend to lack flexibility. Consequently, a more general data fusion scheme is needed. Within this work, a hierarchical Bayesian fusion approach is proposed, which outperforms individual trackers by using redundant measurements. The adaptive framework is achieved by relying on each measurement's local statistics and a global softened majority voting. The proposed approach was validated in a simulated application and two robotic platforms.

\end{abstract}

\section{INTRODUCTION}

In recent years, advancements in visual tracking have allowed the emergence of new robotic platforms capable of following objects with good results. However, robustness is still a major concern within the computer vision community. This is in part due to problems that make it difficult to associate images of a target in consecutive video frames in an unknown scenario. These problems include: motion of the object and/or camera, orientation and pose change, illumination variation, occlusion, scale change, clutter, and the presence of similar objects in the scene. These common disturbances make tracking with any single approach unreliable in many short term scenarios and nearly impossible in most long term applications. While a specific algorithm could work for certain scenarios, it might not work for others. Based on this paradigm, this paper proposes a general tracking approach that fuses the results generated by several algorithms into a unique output. Fusion is done at the bounding box level, where measurements provided by each of the individual tracking algorithms are processed as sensor measurements.

In the literature, sensor fusion is also known as multi-sensor data fusion, data fusion, or combination of multi-sensor information. All of these methods aim for the same goal of creating a synergy of information from several sources \cite{Refworks:Gu}. Normally, observations performed by individual sensors suffer from inaccuracies. A system with only one sensor that observes a physical phenomenon generally cannot reduce its uncertainty without relying on additional sensors. Furthermore, the failure of a sensor leads to a failure of the system as a whole. Different types of sensors provide a spectrum of information with varying accuracy levels and abilities to operate under different conditions \cite{Refworks:Rahimi}. 

There are a number of benefits to data fusion. First, with redundant information, the uncertainty can be reduced to increase the overall accuracy of the system. Second, if a sensor is deemed to be faulty, another sensor might compensate for that fault. Furthermore, while one algorithm could be more robust, say, to scale changes, another could be more robust to outlying measurements; a cooperative approach incorporates the best aspects of each method.

\section{RELATED WORK}

This section describes the different sensor fusion and adaptive sensor fusion  approaches, from general algorithms tailored for fusing sensor measurements to more specific algorithms used in computer vision available in the literature. This overview covers some of the latest sensor fusion mechanisms mentioned in \cite{RefWorks:186}, computer vision benchmarks such as \cite{RefWorks:213} and the performance evaluation of some vision-based trackers \cite{RefWorks:211}.

Initial ideas of adaptive data fusion began in the 1960s \cite{RefWorks:193}, but it was not until the early 1990s that the concept of fusion started to be fully explored \cite{RefWorks:189}, laying the foundation for adaptive Bayesian approaches using the Kalman filter (KF) and its variations based on fuzzy logic \cite{Refworks:Gu,RefWorks:208,RefWorks:209,Refworks:Bostanci} such as the more recent Unscented Kalman Filter (UKF) that uses multiple fading factors-based gain correction \cite{RefWorks:216}. With the recent growth in computational performance, more robust approaches based on the Particle filter (PF) began to emerge \cite{RefWorks:190}. However, both KFs and PFs are known to be susceptible to outliers, and recent studies have tried to solve this problem by introducing extra mechanisms to improve overall robustness \cite{RefWorks:207,RefWorks:200,RefWorks:204}. More complex and time consuming algorithms have gone further by considering not only outliers, but also the type of sensor fault in order to resolve this shortcoming \cite{RefWorks:241}. Additionally, when compared to KFs, PFs are computationally demanding as they tend to require a large number of particles for improved robustness. For this reason, they are not popular in applications that involve moderately high dimensional state spaces.

An adaptive fusion approach with a hierarchical architecture was recently proposed that not only adapts but also encodes information from the performance of the sensors \cite{RefWorks:188}. Although this approach is widely used for model regression and classification, training could leave unexplored regions, causing the resulting output to suffer from outlying data. In addition, depending on the selection of experts, the gating network and the inference model, the overall system cannot be applied in real time applications \cite{RefWorks:187}. 

While adaptive data fusion has been well studied and established for multi-sensor measurements in general \cite{Refworks:drone}, researchers in the computer vision community have gone towards machine learning techniques to incorporate multiple image characteristics into tracking algorithms. Methods such as PROST \cite{RefWorks:195}, VTD \cite{RefWorks:196}, CMT \cite{RefWorks:109}, Struck \cite{RefWorks:141}, or the well known TLD \cite{RefWorks:198} and its variants \cite{RefWorks:121,RefWorks:126} fit this framework. However, the aforementioned algorithms provide limited mechanisms to incorporate multiple and complementary feature extraction methods, thereby restricting their practical applicability. 

Some of the latest visual tracking fusion approaches suggest fusion at the bounding box level \cite{RefWorks:194}, where information such as pixel coordinates are readily available. However, to achieve such fusion, offline training and weight finding must be carried out. This is achieved using ground truth (GT) information as well as performance metrics of the dataset used to train the algorithms. More general fusion approaches have been recently proposed, most of which rely on Sequential Monte Carlo Bayesian methods such as  PFs \cite{RefWorks:220,RefWorks:219,RefWorks:235}, and are hence too computationally demanding for real-time control applications.  

This work aims to create a general Bayesian approach for real-time applications in robotic platforms. The proposed method processes the bounding boxes of the trackers/detectors as sensor measurements. This framework is founded in the basis of the bank of KFs with some similarities with the mixture of experts aforecited. Furthermore, this scheme addresses some common problems such as data imperfection, outliers and spurious data, measurement delays, static vs. dynamic phenomena, and others discussed in \cite{RefWorks:186}. Additionally, this approach was tested in simulated signals and two different robotics platforms: An UAV system and a pan-tilt system. Both are capable of following a target. While similar approaches have used vision-based trackers to control a small UAV in  \cite{RefWorks:143,Refworks:Changhong} and \cite{RefWorks:126}. Previous works did not consider the fusion of several methods at a bounding box level to improve reliability over longer time spans.

\section{System description}
To avoid confusion, all visual trackers/detectors used in this work that produce a bounding box such as DSSTtld \cite{RefWorks:126}, CMT \cite{RefWorks:109}, or Struck \cite{RefWorks:141} will be called detectors from this point forward. These algorithms are processed as sensors that cast measurements. The method proposed in this work, which we call Hierarchical Adaptive Bayesian Data Fusion (HAB-DF), is the main tracker that processes such measurements. 

\begin{figure}[!ht]
\centering
\includegraphics[width=0.25\textwidth]{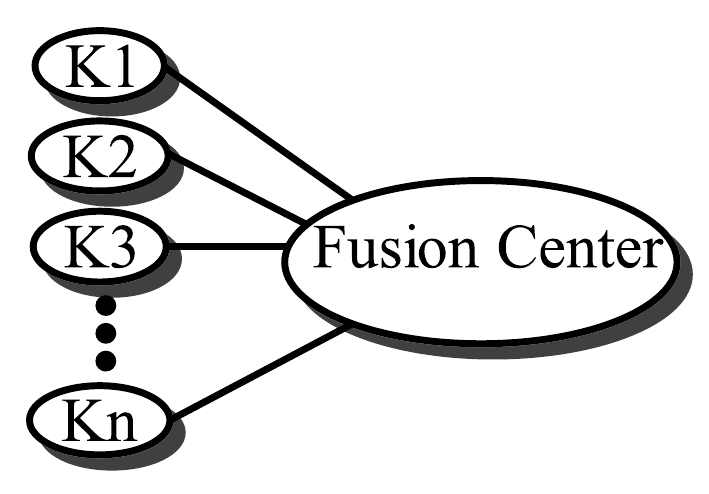}
\caption{Hierarchical Adaptive Bayesian Data Fusion approach. The first level of the hierarchy consists of experts that provide a local estimate to the fusion center. The second level is the fusion center. }
\label{fig:K_expertes}
\end{figure}

The approach proposed in this paper is a variation of the framework commonly known as mixture of experts \cite{RefWorks:187}, which are organized in levels or hierarchies that converge in a gating network. This work substitutes that gating network with a Bayesian approach that adapts online. Therefore, no training is necessary. In addition, this method is organized in two levels or hierarchies: the experts and the fusion center. Each expert module, $K_i$, $i=1,...n$, works asynchronously from the other modules. Usually, a bank of estimators is applied when the sensors differ in model, as each suffers from different failure types. In this particular case, the experts are KFs, inspired in part by \cite{RefWorks:218} and \cite{RefWorks:188}. Figure~\ref{fig:K_expertes} shows a representation of this approach. 

In the hierarchical model, each expert is equipped with an outlier detection mechanism that calculates a reliability score. The fusion center merges the outputs of each expert by adopting a weighted majority voting scheme.  

\subsection{Bayesian State Space Model }

A general model of  a linear Kalman Filter derived from \cite{RefWorks:124} is used. The state vector is given by $\textbf{x}$ = $[u$ $v$ $h$  $w$ $\dot{u}$ $\dot{v}$ $\dot{h}$ $\dot{w}]$, where $u$, $v$ are the pixel coordinates of the center of the target, $h$ and  $w$ are its height and width, respectively. $\dot{u}$ $\dot{v}$ $\dot{h}$ $\dot{w}$ are the velocities in each dimension. In this work, we adopt a random acceleration model. The object tracking system is then represented as follows:
% state dynamics
\begin{align} \label{eq:state dynamics}
\textbf{x}(t) &= A\textbf{x}(t-1) + B\textbf{u}(t) + \textbf{w}(t) & \\
\textbf{y}(t) &= C\textbf{x}(t) + \textbf{v}(t) \label{eq:observation model}
\end{align}
where Eq.~\eqref{eq:state dynamics} represents the system dynamics, including the state transition matrix $A$, the influence of the control action $B$ and the process noise $\textbf{w}$. Eq.~\eqref{eq:observation model} is the measurement model, which includes the observation matrix $C$ and the measurement noise $\textbf{v}$. The process noise and measurement noise are assumed to be white and Gaussian, with variances $R_{ww}$ and  $R_{vv}$ respectively. That is, \newcommand\pN{\mathcal{N}} $\textbf{w} \sim \pN(0, R_{ww})$ and $\textbf{v} \sim \pN(0, R_{vv})$.

\subsection{Hierarchical Adaptive Bayesian Data Fusion}

To reduce the sensor fusion uncertainty, two approaches have been implemented. One approach is concerned with the reliability of the measurement, delivering a local estimate based on the Mahalanobis distance \cite{RefWorks:192}. The other is a global approach based on majority voting. The overall approach is divided into a two-level hierarchy: experts and the fusion center. While each expert uses position and speed for accuracy, the fusion center only fuses direct measurements such as position, but still predicts speeds for better results in subsequent frames. Furthermore, this concept is not limited to KFs. Any Bayesian estimator can be used to accomplish fusion. Nevertheless, KFs are known for being efficient, fast, and ideal for real-time applications. 

\subsection{Local Expert Weighting}

Like other filters, KFs are susceptible to abnormally large error in estimation. This is in part due to  KFs not being robust to outliers. Several works have been proposed to solve this dilemma \cite{RefWorks:200,RefWorks:202,RefWorks:203}. The Mahalanobis distance (MD) alleviates this issue by providing a measure of how much a predicted value differs from its expected distribution.

Outliers occur due to modeling uncertainties, incorrect process/measurement noise covariances selection, and other external disturbances.  If the estimation error (the difference between the real state and the estimated state) of the KF is beyond a certain threshold, the MD can penalize the expert as being in failure or abnormal mode. Alternatively, one can use the predicted measurement to determine outliers. This error is then defined as follows: given a measurement $\textbf{y} =$ $[y_1$ $y_2$ $...$ $y_N]^{T}$, the MD from this measurement to a group of predicted values with mean
$\boldsymbol{\mu}	 =$ $[\mu_{1}$ $\mu_{2}$ $...$ $\mu_{N}]^{T}$ and covariance matrix $C$ is given by
\begin{equation} \label{eq:m_dist}
M(\textbf{y}) = \sqrt{(\textbf{y}-\boldsymbol{\mu})^{T}C^{-1}(\textbf{y}-\boldsymbol{\mu})}
\end{equation}
Since each expert is equipped with its own MD calculation, an approximated version is used \cite{RefWorks:147}:
\begin{equation} \label{eq:m_dist_aprox}
M({y}) \approx \sum_{i=1}^{N} \left(\frac{{q_{i}}^2}{C_{i}}\right)^{1/2}
\end{equation}
where ${q_{i}=y_{i}-\mu_{i}}$ and $C_{i}$ is the $i^{th}$ value along the diagonal of the innovation covariance $C$. Eq.~\eqref{eq:m_dist_aprox} decreases the computational burden if a considerable number of experts is needed. Usually, an estimator can be penalized if the MD is beyond a certain threshold. However, doing so yields hard transitions. To soften this rule, a sigmoid function has been employed ~\cite{Guevara2016Vision}:
\begin{equation} \label{eq:w_MD}
w_{M}=\frac{1}{1+e^{(-M(\textbf{y})+\xi)}}
\end{equation}
where $\xi$ is a value chosen using the $\chi^2$ distribution based on the number of degrees of freedom (DOF) of the system and the desired confidence level. Outliers are identified using Eq.~\eqref{eq:w_MD} where $w_{M}$ represents an expert's performance in the form of a local weighting function. 

\subsection{Majority Voting}

Voting is one of the simplest approaches for fusing information \cite{RefWorks:193}. There are many ways to determine the weights in a majority voting scheme. The method chosen for this application is a weighted decision that combines the output of multiple sensors (in this case, information from multiple bounding boxes).

\begin{figure}[!h]
\centering
\includegraphics[width=0.3\textwidth]{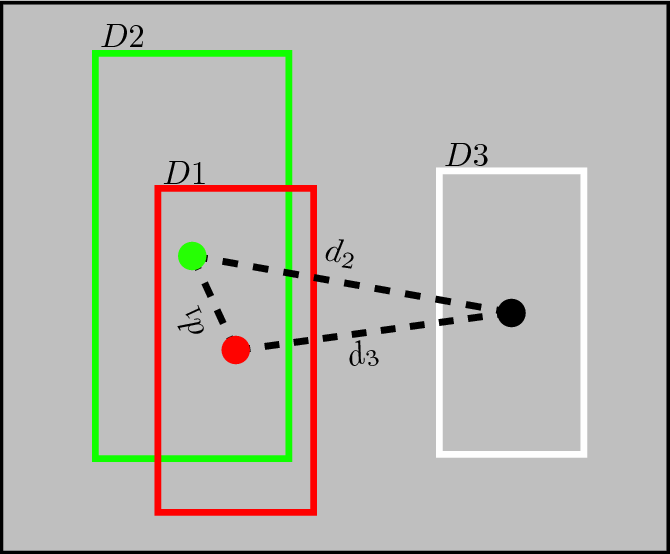}
\caption{Majority voting representation. Distances $d_i$ are traced from the center of each detector. While these distances are shown as the center distances among detectors ($u$ and $v$), they also comprise their heights and widths ($h$ and $w$). In this scenario, $D_1$ and $D_2$ are close to each other, while $D_3$ is farther away. The consensus will penalize $D_3$ in this case, since $d_1$ is the minimum distance.}
\label{fig:BB_repr}
\end{figure}

This method begins by calculating the pairwise Euclidean distance between bounding boxes 
\begin{equation} \label{eq:ED_C}
\begin{split}
\MoveEqLeft
d_i{(\textbf{p},\textbf{r})}=\norm{\textbf{p}-\textbf{r}} \\ 
&i=1,2,3,\cdots,n 
\end{split}
\end{equation} 
where $\textbf{p}$ and $\textbf{r}$ are vectors that represent the coordinates and the size of the bounding boxes for two different detectors $D_i$ and $D_j$. A statistical descriptor such as the minimum value can be used to reach consensus among all the detectors
\begin{equation} \label{eq:min_desc}
\begin{split}
\MoveEqLeft
min_{d}=\min(d_{i}, \cdots, d_{n})\\
&i=1,2,3,\cdots,n 
\end{split}
\end{equation} 
Figure~\ref{fig:BB_repr} shows a scenario in which detector $D_3$ would be penalized because it is farther from the other two detectors. Note that this scheme imposes no limit to the number of detectors/sensors that can be used. The only limitation is computational performance. Although a minimum of 3 detectors/sensors is needed so that a consensus can be reached.

To calculate a weight that penalizes detectors for being farther from the cluster of detectors, instead of using a hard limiter, a hyperbolic tangent is applied, allowing a soft transition among detectors:
\begin{equation} \label{eq:w_desc}
\begin{split}
\MoveEqLeft
w_d=\omega_0+\omega(1+\tanh(min_{d}-\lambda))
\end{split}
\end{equation} 
where $\omega_0$ is an initial weight consistent with the observed phenomenon, $\omega$ is the desired impact of the penalization function, which determines the overall effect of a particular detector in the fusion if it drifts away, and $\lambda$ determines the distance at which the penalization starts taking place.

\subsection{Adaptive Fusion Center Strategy}

The bank of KFs is composed of one filter for each sensor/detector. Each filter/expert in the bank gives a local estimate of the detector/measurement assigned to that particular filter. Another KF acts as the fusion center, which adapts itself at each measurement by updating its measurement noise covariance according to 
\begin{equation} \label{eq:COV_MAT}
R_{vv}(w_{d},w_{M})= \Gamma w_{d}+\Delta w_{M}
\end{equation} 
where $w_d$ and $w_M$ are given by Eqs. \ref{eq:w_desc} and \ref{eq:w_MD}, respectively, $\Gamma= diag(\gamma_1,\gamma_2,\cdots,\gamma_n)$, $\Delta= diag(\delta_1,\delta_2,\cdots,\delta_n)$, and $diag(.)$ represents a diagonal matrix whose elements are the function parameters. $\gamma_i$ and $\delta_i$ can be set to 1 if there is no a priori knowledge of the system. Otherwise, $\gamma_i$ can be set  to a  value depending on the knowledge of the noise of the sensor and $\delta_i$ can be set to a value depending on how much drift the sensor suffers.

\section{Platform description}

A pan-tilt system and a small UAV were used to test the proposed method. The algorithm was implemented in C++ and ran in a Lenovo W530 laptop with an Intel® Core™ i7-3630QM CPU @ $2.40GHz \times 8$ processor and a Quadro K1000M graphics card.

\subsection{Pan-Tilt System}
\label{sec:pts}

The platform was composed of two servo motors that control the $2DOF$ of the system with an on-board Creative Senz3D camera\footnote{Only RGB  images were used in this work. Depth data was discarded.}. Two different PID controllers kept the system as close as possible to the center of the image by using the centroid of the fusion approach. The servo motors were driven by the computer using an \textit{Arduino UNO} that converted the position commands into PWM signals for the servo motors. Position commands were sent using serial communication. The implemented PID gains for both the pan and tilt motions were: $Kp=35$, $Ki=3.4$ and $Kd=8$.

\subsection{UAV Platform}
\label{sec:uavp}

The UAV used in this work was the Parrot AR.Drone 2.0, controlled over a Wi-Fi link. The $4DOF$ platform is controlled using the same heuristic proposed in \cite{RefWorks:143}. However only a PD controller was used, with the following gains:
\begin{itemize}
\item Pitch($\theta$): $Kp_{\theta}=0.020$ and $Kd_{\theta}=0.020$.
\item Roll($\phi$): $Kp_{\phi}=0.699$ and $Kd_{\phi}=0.400$.
\item Yaw($\psi$): $Kp_{\psi}=0.120$ and $Kd_{\psi}=0.020$.
\item Throttle: $Kp_T=0.430$ and $Kd_T=0.021$.
\end{itemize}
Furthermore, in addition to attempting to keep the target at the center of the image using its centroid position ($u$,$v$), the UAV also used the target's relative scale variations, based on $h$ and $w$, to keep a constant distance from the target.

\section{Experimental Results}

Several experiments were conducted to evaluate the proposed HAB-DF approach, from a simulation-based experiment to real applications using the pan-tilt system and the UAV platform described in Sections \ref{sec:pts} and \ref{sec:uavp}.

\subsection{Simulations}

\begin{figure}[!h]
    \begin{center} 
    \begin{subfigure}[h]{.65\columnwidth}
        \includegraphics[width=\columnwidth]{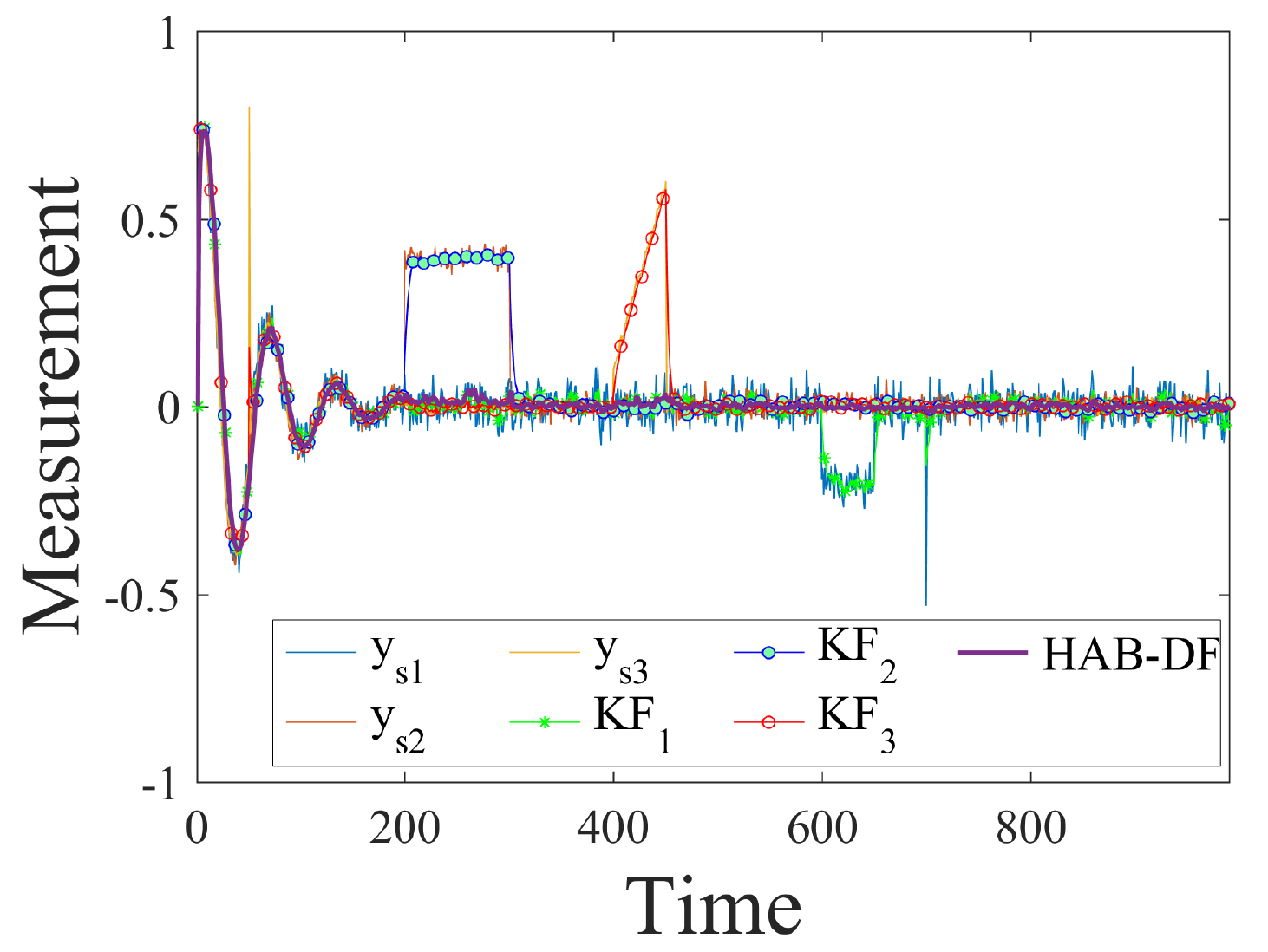}
        \caption{Simulated signals}
        \label{fig:Sim_SI}
    \end{subfigure}%
    \end{center}
    
    \begin{center} 
    \begin{subfigure}[h]{.65\columnwidth}
        \includegraphics[width=\columnwidth]{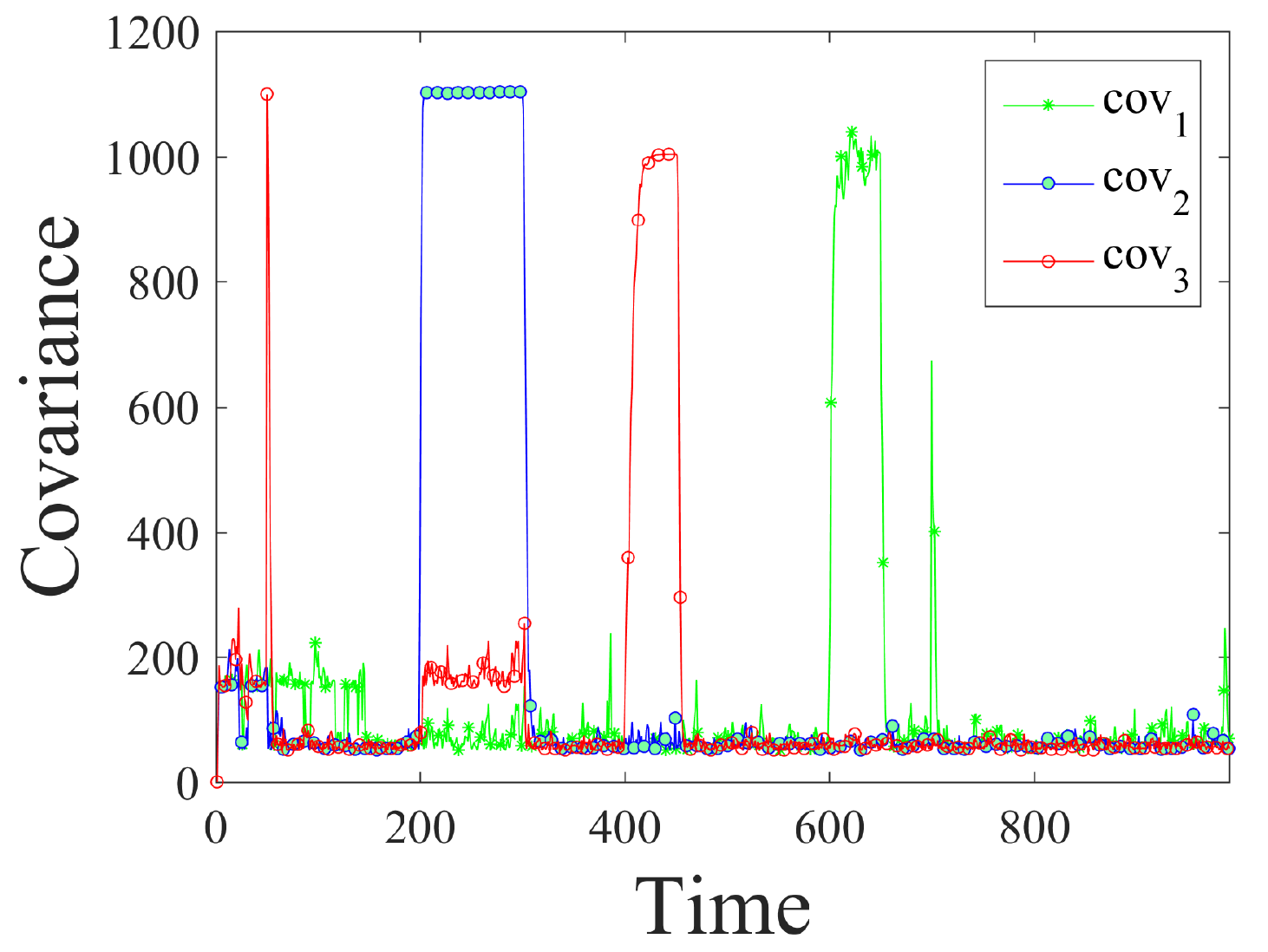}
        \caption{Adaptive Covariance}
        \label{fig:Sim_COV}
    \end{subfigure}     
    \end{center} 
\caption{Simulation of a second order system. $y_{si}$, $i=1,2,3$ are the sensor measurements. The HAB-DF is the only method that is able to accurately track the signal by fusing the output of each KF in the first hierarchy. Each sensor suffers from different types of faults: Gaussian noise, spikes, drifts and shocks (a constant offset for an given time). }
    \label{fig:Simulation}
\end{figure}

A simulation using the HAB-DF is shown in Figure~\ref{fig:Simulation}. To emulate a scenario in which different sensors have distinct characteristics, each signal in the simulation suffers from different types of noise and faults. Each expert in the first level of the hierarchy fed the fusion center with its own estimate. Having redundancy in sensor data produced estimations that no single method could accomplish alone. Moreover, the way that the approach adapts itself along the run allows it to eliminate noise and faults. This can be seen in Figure~\ref{fig:Sim_COV}, where higher covariance values indicate that each expert in the first hierarchy is deemed faulty depending on its performance.

Compared to others works like \cite{RefWorks:241}, the HAB-DF took outliers into consideration by using the Mahalanobis distance, softening their impact. Unlike \cite{RefWorks:241},  HAB-DF does not learn the fault types, as learning specific types can leave unexplored regions outside the scope of the training scenarios. Additionally, the majority voting penalizes any faulty sensor.

 \setcounter{figure}{4}

\begin{figure*}[!b]
\centering
\begin{subfigure}[t]{.65\columnwidth}
\includegraphics[width=\columnwidth]{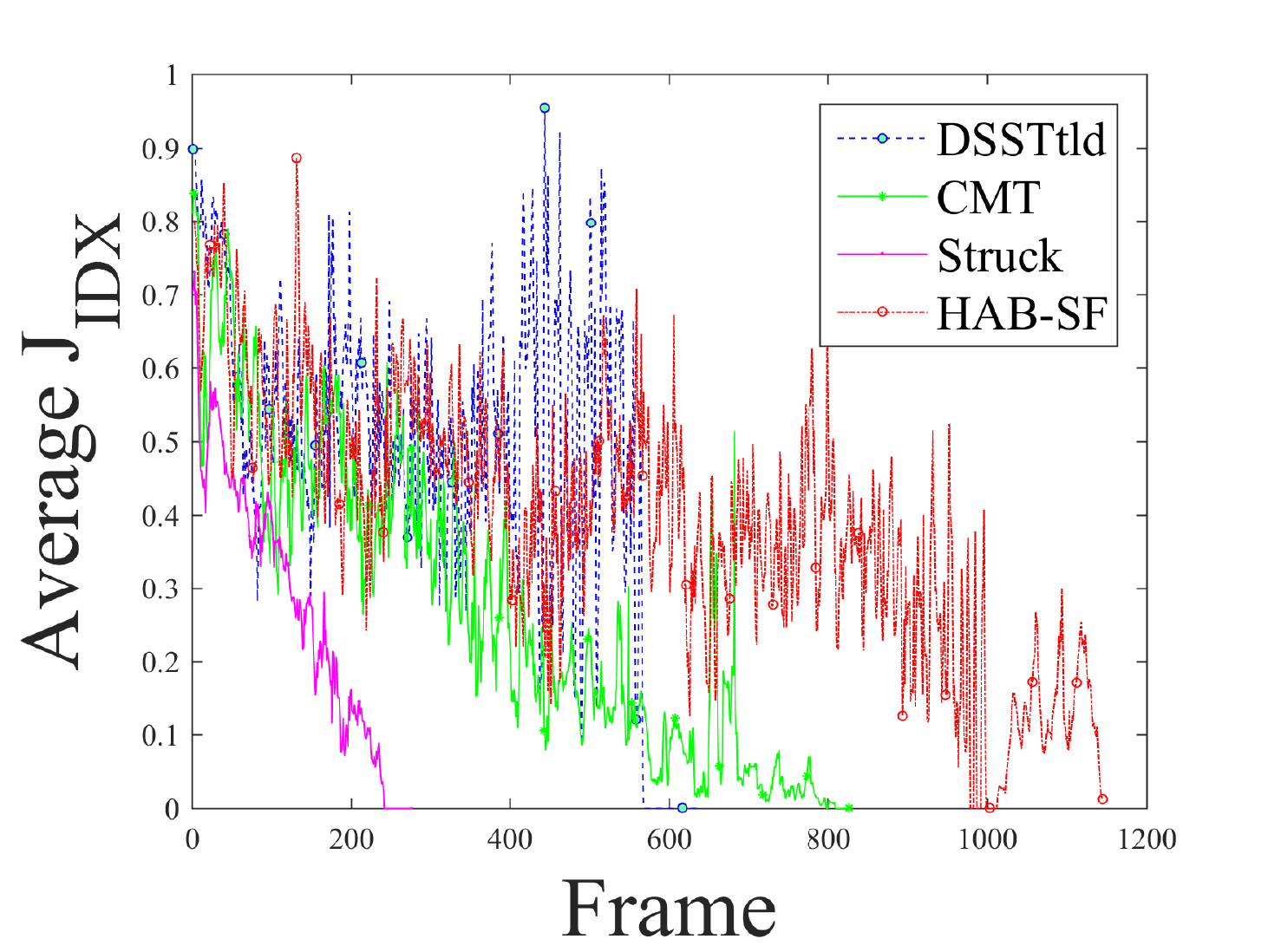}%
        \caption{Jaccard Index}
        \label{fig:Jc_index}
\end{subfigure}
~
\begin{subfigure}[t]{.65\columnwidth}
\includegraphics[width=\columnwidth]{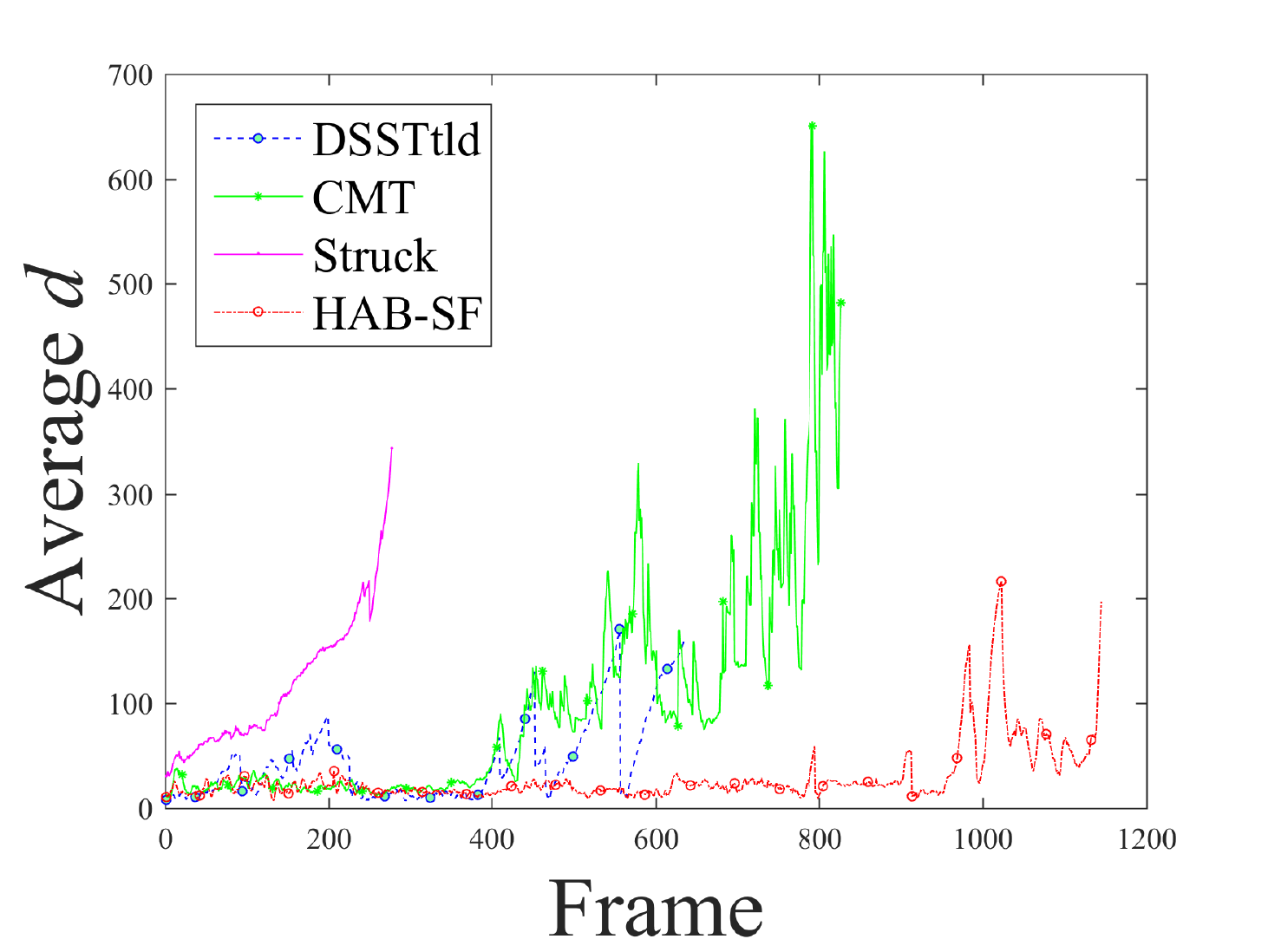}%
        \caption{Euclidean distance $d$}
        \label{fig:4D_performance}
\end{subfigure}
~
\begin{subfigure}[t]{.65\columnwidth}
\includegraphics[width=\columnwidth]{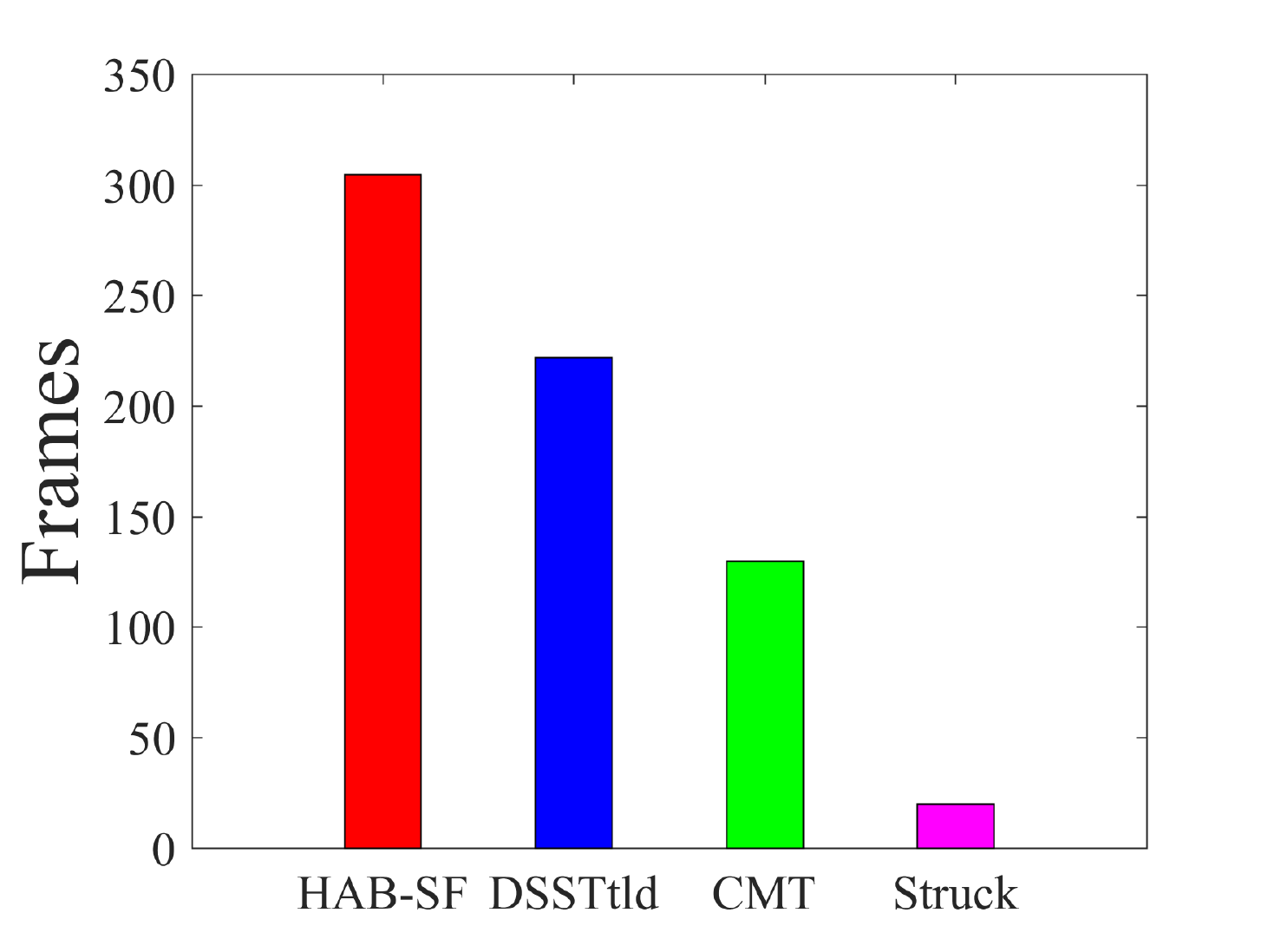}%
       \caption{Measure of success}
        \label{fig:com_per}
\end{subfigure}%
\caption{Average performance. Figure~\ref{fig:Jc_index} shows the average of the $J_{IDX}$. A decrease in $J_{IDX}$ indicates a tracking performance degradation. A value of zero indicates a complete failure in which there is no overlap between the GT and the detector. Figure~\ref{fig:4D_performance} shows average  of $d$ for each approach. A value close to zero means that the GT and the tracker are similar. Figure~\ref{fig:com_per} shows the success bar graph, a frame is considered successfully tracked when $J_{IDX} \geq 0.5  \:\:\:\: and \:\: \:\: d \leq 50$.  }
\label{fig:av_perf}
\end{figure*}

\subsection{Pan-tilt System}

This section describes the experiments carried out using the pan-tilt platform presented in Section \ref{sec:pts}. The evaluation consisted of testing each of the estimators individually with their respective detector and then the fusion of all of the detectors. This experiment took place in a room where a face was tracked using the pan-tilt system. All the experiments were run using similar light conditions and with the same face at similar starting distances. Each run lasted until the target was out of the image frame or noticeable tracking loss occurred. This gave a result where each individual test follows the target for a different number of frames. Furthermore, to compare each individual detector's overall performance, each test was labeled by hand. Since manual labeling is a time consuming task, it was decided to test each estimator and the proposed HAB-DF for 5 iterations, for a total of 20 data sets. Figure~\ref{fig:Rnd_Img} presents images from these selected sequences.

 \setcounter{figure}{3}
\begin{figure}[!h]
		\medskip
        \begin{subfigure}[!h]{0.23\textwidth}
                \includegraphics[width=\textwidth]{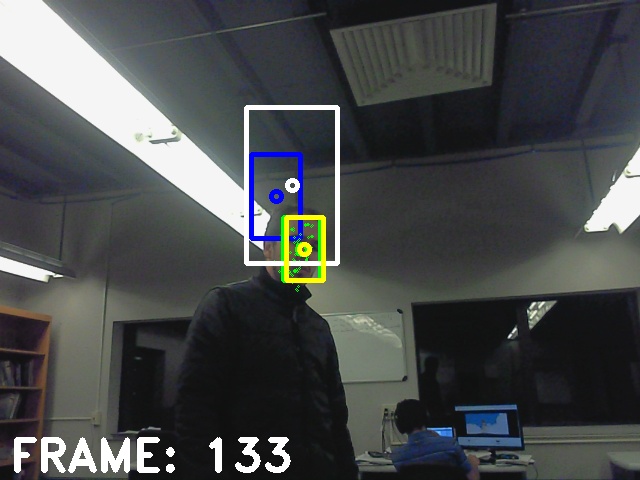}
                \caption{HAB-DF}
        \label{fig:HAB-SF_T}
                
        \end{subfigure}
        \begin{subfigure}[!h]{0.23\textwidth}
                \includegraphics[width=\textwidth]{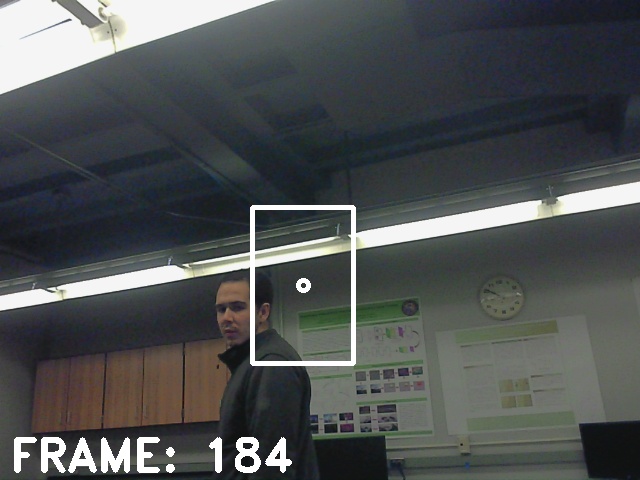}
                \caption{Struck}
        \label{fig:Struck_T}
        \end{subfigure}
        ~
        \begin{subfigure}[!h]{0.23\textwidth}
                \includegraphics[width=\textwidth]{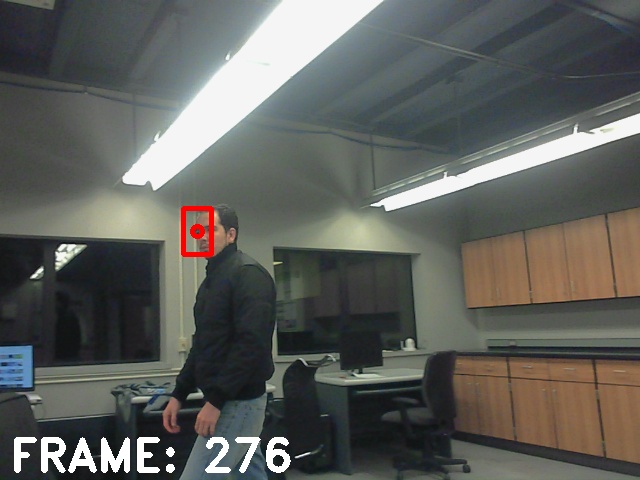}
                \caption{DSSTtld}
        \label{fig:DSStld_T}
        \end{subfigure}
        \begin{subfigure}[!h]{0.23\textwidth}
                \includegraphics[width=\textwidth]{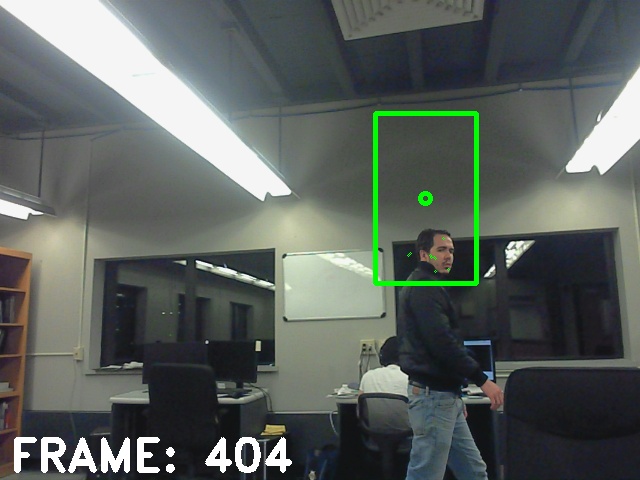}
                \caption{CMT}
        \label{fig:CMT_T}
        \end{subfigure}
               \caption{Pan-tilt system experiment (best seen in colors). The frames shown here are random frames selected from the dataset. Each of them presents a different tracking approach. The target was moving sideways with some vertical disturbances, and gradually increasing the distance from the camera. In (a) HAB-DF is shown in yellow and DDSTtld is shown in blue because it is lost.}
\label{fig:Rnd_Img}
\end{figure}

Performance and reliability were measured with an overlap score (also known as the Jaccard index), given by
\begin{equation} \label{eq:JAC_ind}
J_{IDX}(A_{bb},A_{T})=\frac{A_{bb} \cap A_{T}}{A_{bb} \cup A_{T}} 
\end{equation} 
where $A_{bb}$ and $A_{T}$ are the areas in pixels of the bounding boxes of each approach and of the GT, respectively. $J_{IDX}$ measures the area of overlap between the bounding boxes generated by each approach and the labeled GT. The closer to 1, the better the performance. In addition to the $J_{IDX}$, the Euclidean distance $d$ used in the majority voting also depicts the dissimilarity among each approach and the GT. Calculating this measure involves the center of the bounding box, its height, and its width. 

 \setcounter{figure}{5}
\begin{figure*}[hb]
    \centering
    \begin{subfigure}[h]{.65\columnwidth}
        \centering
        \includegraphics[width=\columnwidth]{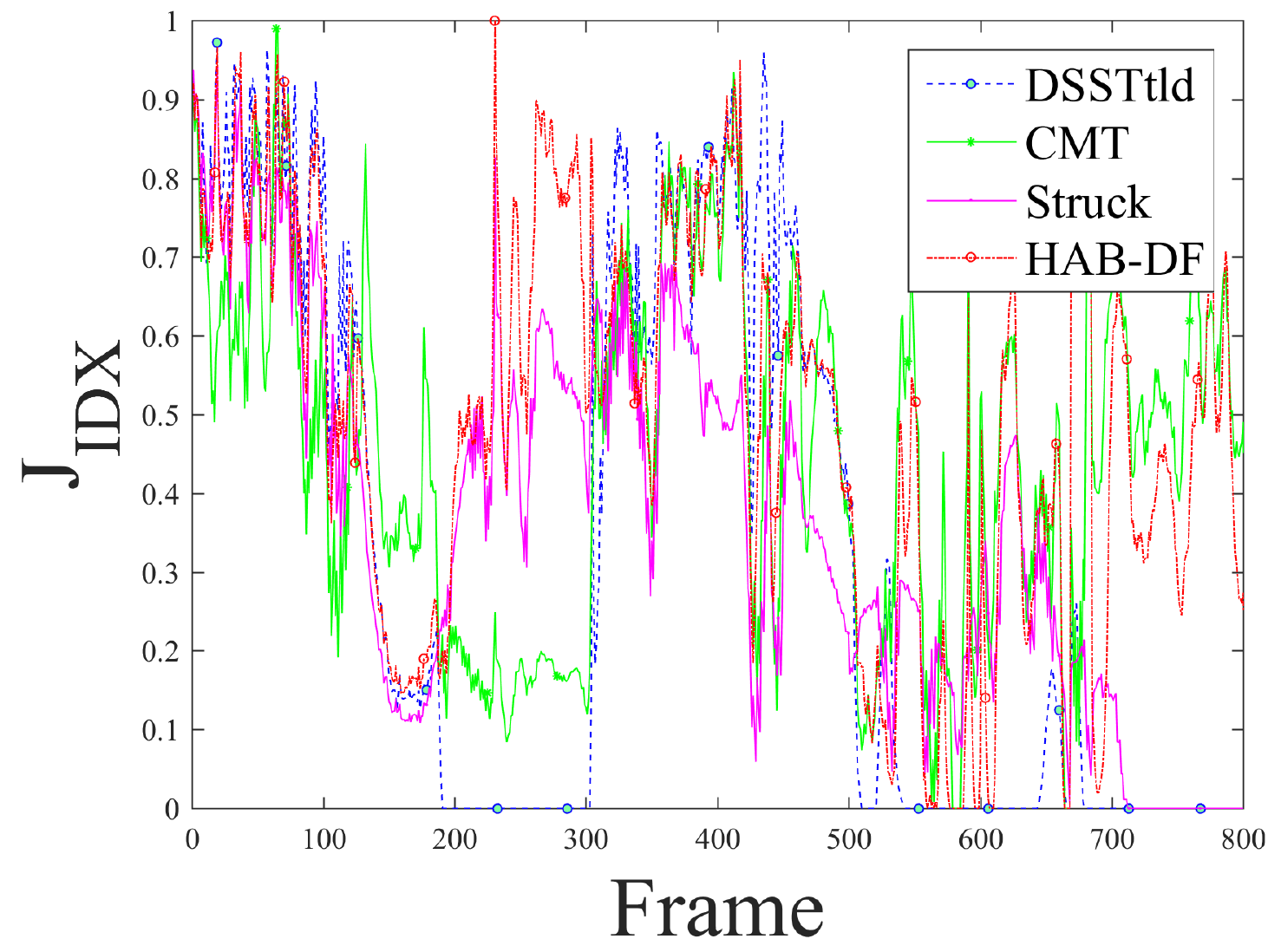}
        \caption{Jaccard Index}
        \label{fig:Jc_ind_11}
    \end{subfigure}%
    ~ 
    \begin{subfigure}[!h]{.65\columnwidth}
        \centering
        \includegraphics[width=\columnwidth]{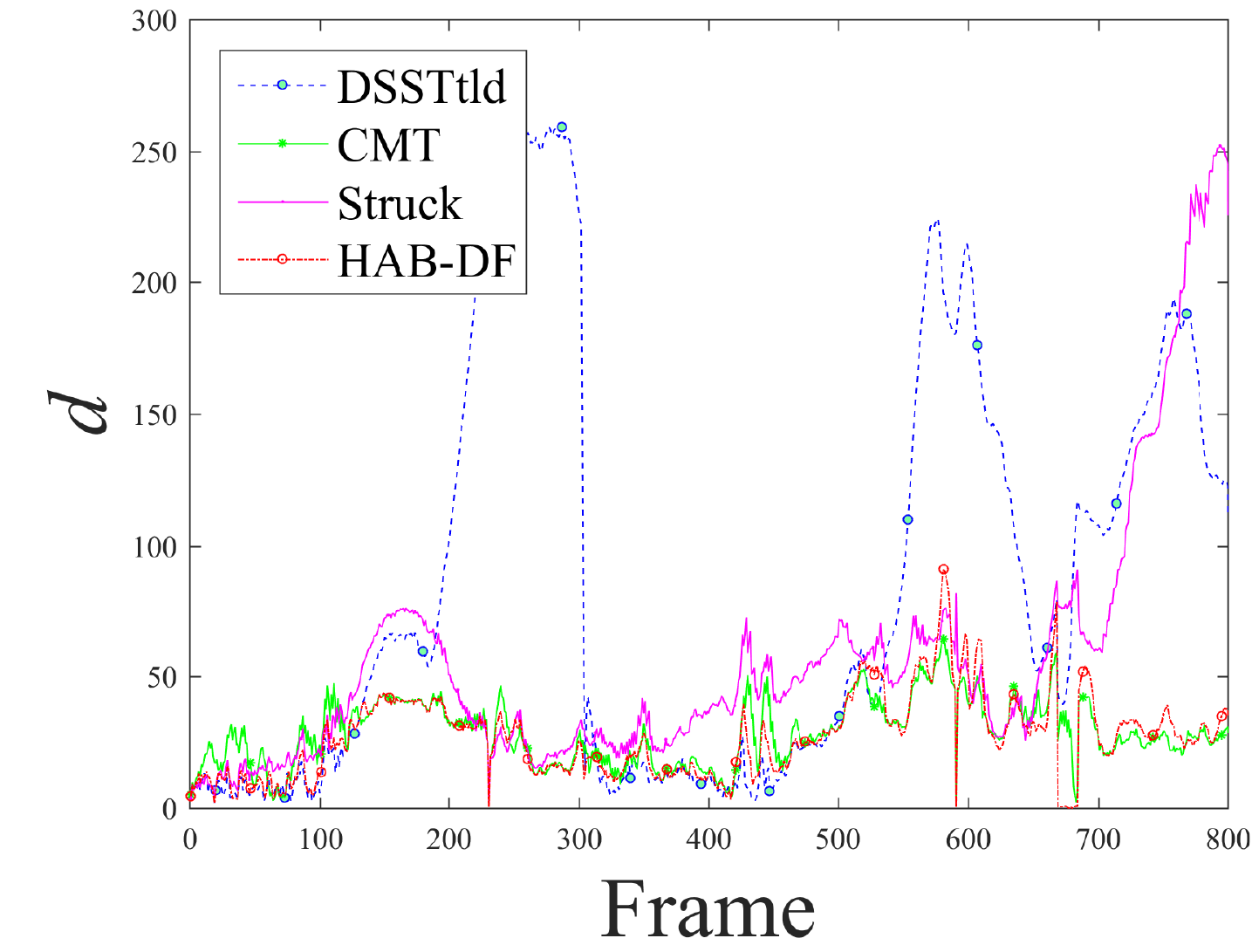}
        \caption{Euclidean distance $d$}
        \label{fig:4D_d_11}
    \end{subfigure}
     ~
    \begin{subfigure}[!h]{.65\columnwidth}
        \centering
        \includegraphics[width=\columnwidth]{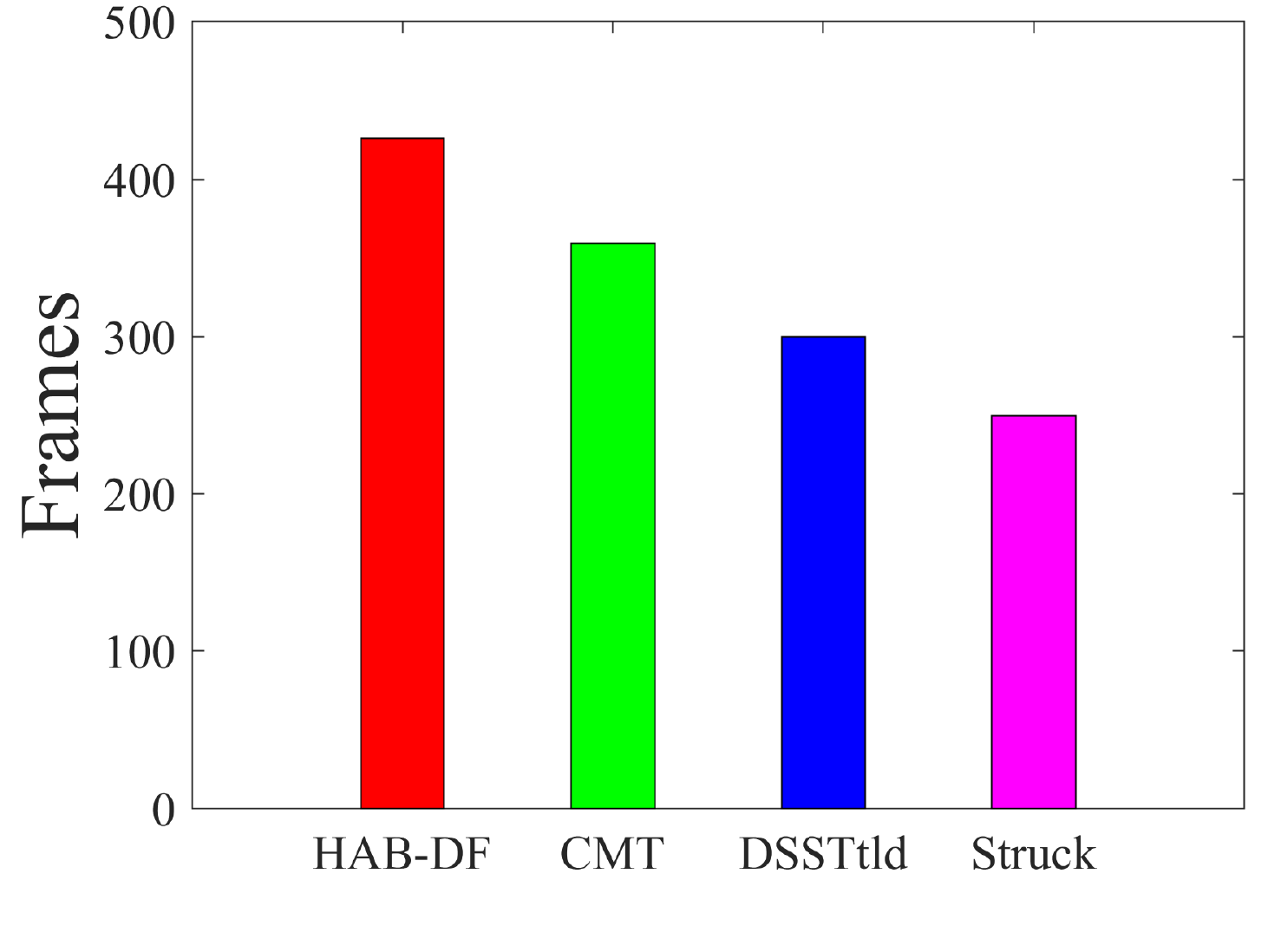}
        \caption{Measure of success}
        \label{fig:succ_bar_11}
    \end{subfigure}
     ~ 
    \begin{subfigure}[!h]{.65\columnwidth}
        \centering
        \includegraphics[width=\columnwidth]{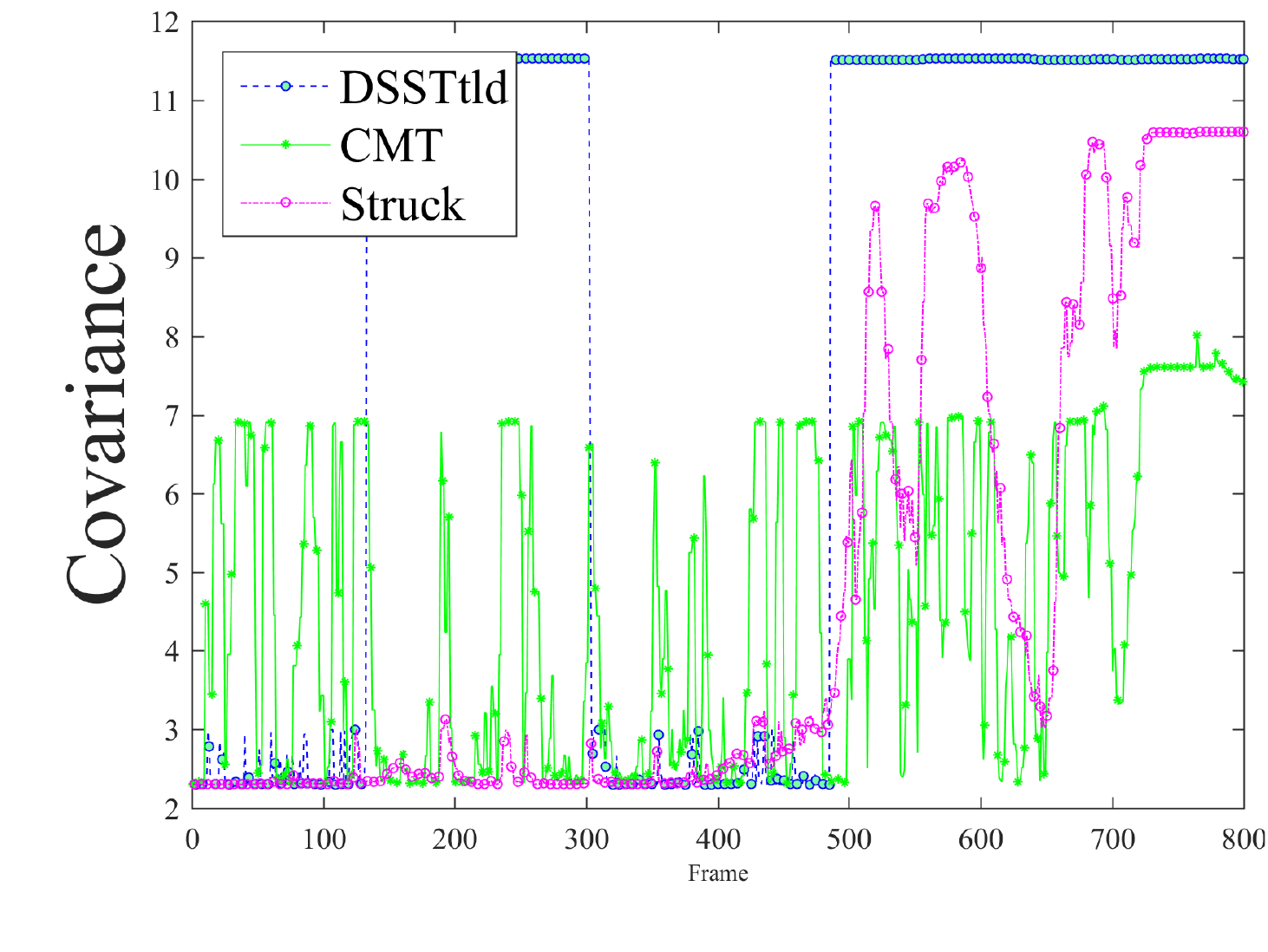}
        \caption{Adaptive Covariance}
        \label{fig:cov_plot_11}
    \end{subfigure}
    ~
    \begin{subfigure}[!h]{.65\columnwidth}
        \centering
        \includegraphics[width=\columnwidth]{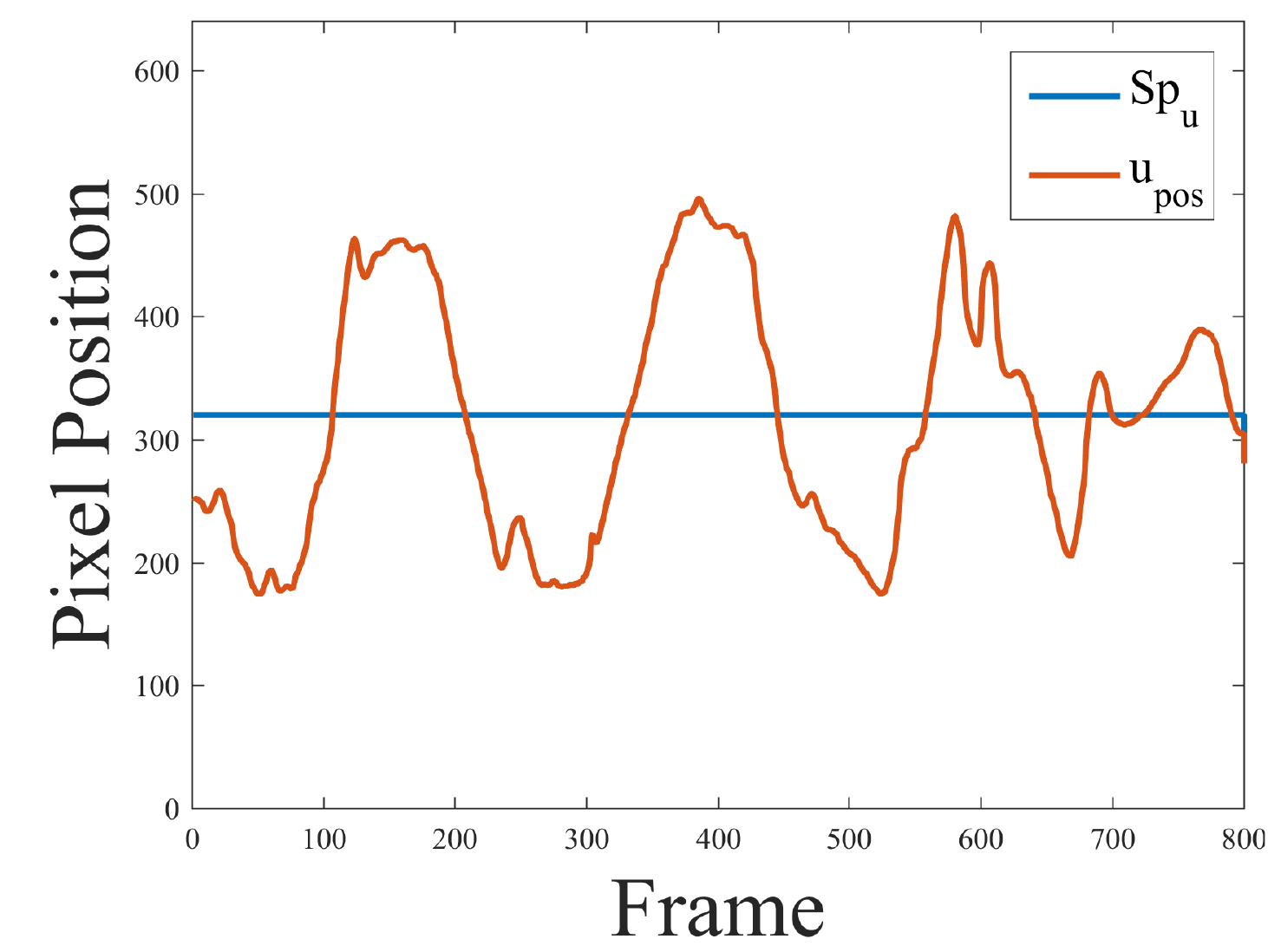}
        \caption{$x$ target position}
        \label{fig:con_spx_11}
        \end{subfigure}
        ~
    \begin{subfigure}[!h]{.65\columnwidth}
        \centering
        \includegraphics[width=\columnwidth]{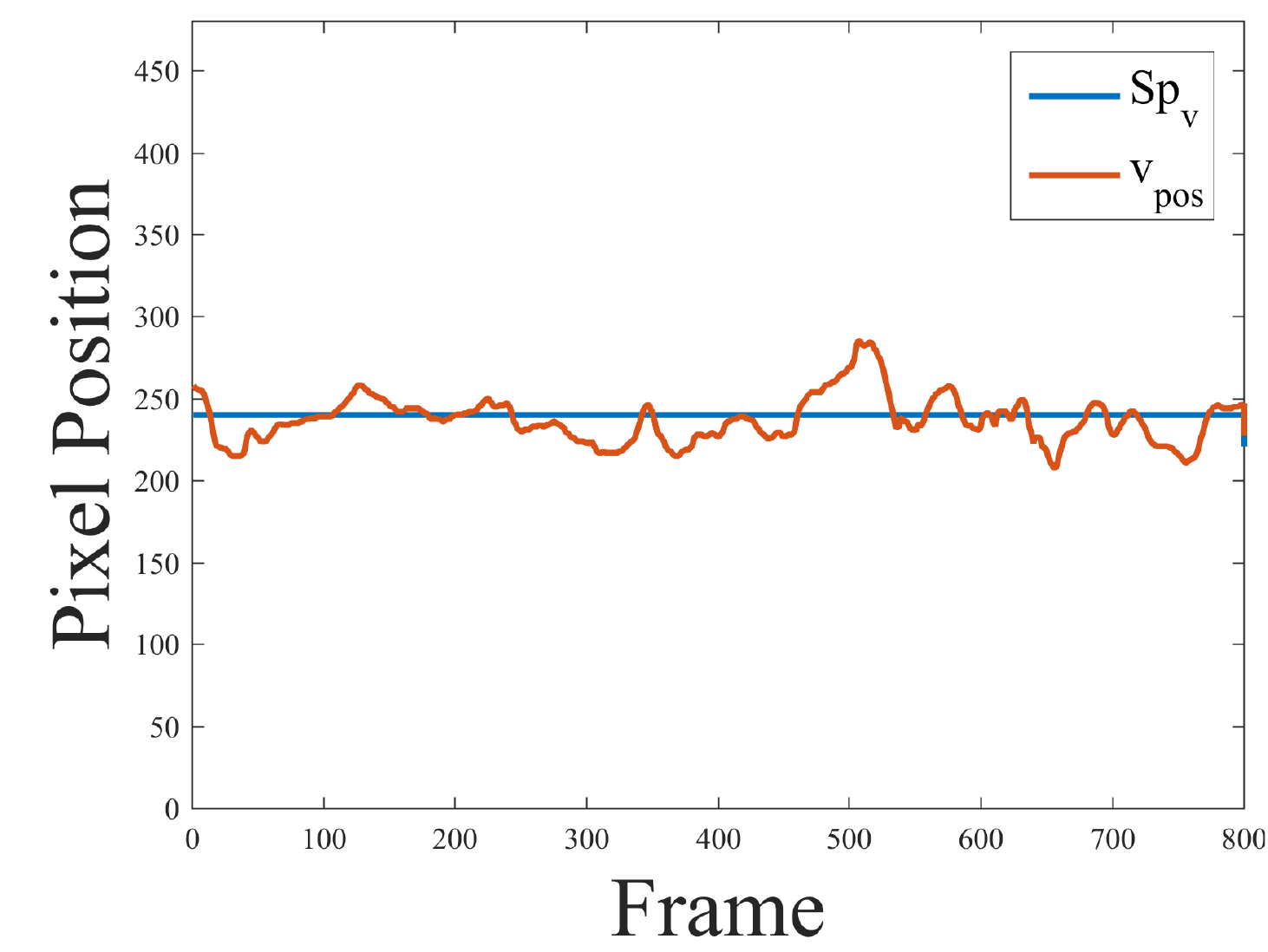}
        \caption{$y$ target position}
        \label{fig:con_spy_11}
        \end{subfigure}
      
\caption{Evaluation of the performance of tracking a recycling bin. Figure~\ref{fig:Jc_ind_11} and Figure~\ref{fig:4D_d_11} show that DSSTtld has a degraded performance (around frames 100-300 and 500-800). This is consistent with Figure~\ref{fig:cov_plot_11}, where DSSTtld suffers of a sudden drop of confidence value resulting in an increment of the covariance that is ruled by the MD and the majority voting scheme. The HAB-DF has the best performance among all the approaches as seen in Figure~\ref{fig:succ_bar_11}. Moreover, the transition between detectors is soft, allowing for the smooth motion control that can be seen in Figure~\ref{fig:con_spx_11} and Figure~\ref{fig:con_spy_11}. }
    \label{fig:trash_bin}
\end{figure*}

Figure~\ref{fig:av_perf} displays several metrics that illustrate the performance of the approaches.  Figure~\ref{fig:Jc_index} shows the average performance of the different detectors and the proposed approach according to $J_{IDX}$. As shown, Struck performed worst among all the detectors, having  problems with scale changes caused by the target moving closer and farther from the camera. CMT, DSSTtld and the proposed approach performed similarly until the $400^{th}$ frame. DSSTtld showed the best performance for a few frames in terms of accuracy (between the $400^{th}$ and $600^{th}$ frame) but was not able to handle pose changes nor out-of-plane rotations of the target, which resulted in a sudden drop in confidence level and consequently losing track of the target. While CMT was able to handle distortions caused by rotation, its $J_{IDX}$ degraded with scale changes. As a result, it kept track of the target longer than the other detectors, albeit with substantially reduced accuracy. If the intrinsic properties  of the detectors are combined, the Bayesian approach is not only more robust but also more accurate than only using a single detector. Also, if one of the detectors is not performing well, such as Struck in the aforementioned scenario, it is possible to see that the fusion is not affected due to its low reliability. Figure~\ref{fig:com_per} shows a comparison of the accuracies of the different approaches. This plot considers a threshold between $J_{IDX}$ and $d$ of what is considered a successful frame. On average, the Bayesian fusion yielded better results and outperformed every single estimator.

\begin{figure}[!t]
		\medskip
        \centering
        \begin{subfigure}[!h]{0.23\textwidth}
                \includegraphics[width=\textwidth]{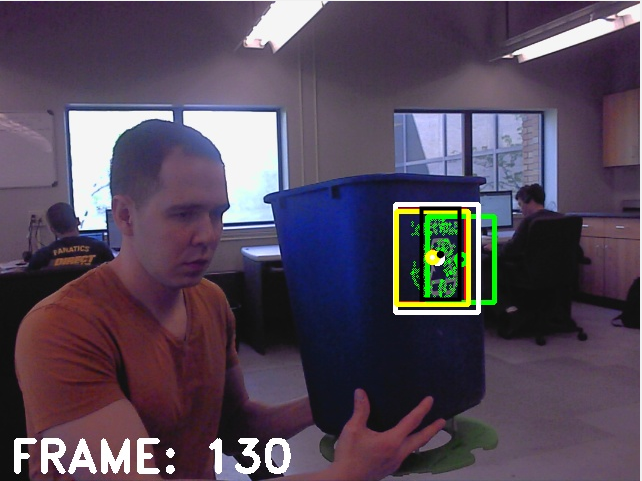}
                \vspace{-1em}
        \end{subfigure}
        \begin{subfigure}[!h]{0.23\textwidth}
                \includegraphics[width=\textwidth]{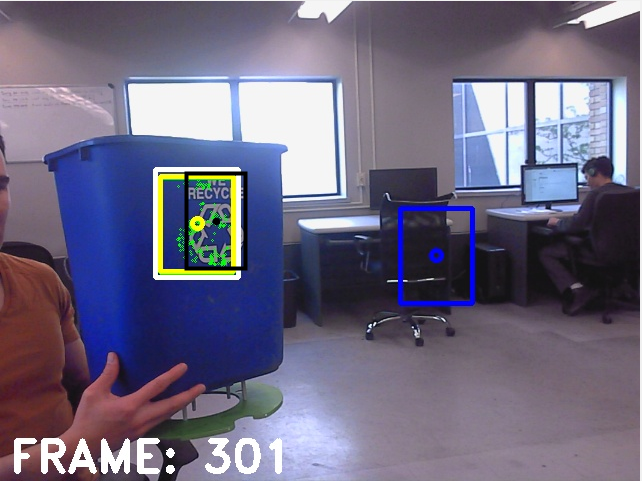}
                \vspace{-1em}
        \end{subfigure}
        ~
        \begin{subfigure}[!h]{0.23\textwidth}
                \includegraphics[width=\textwidth]{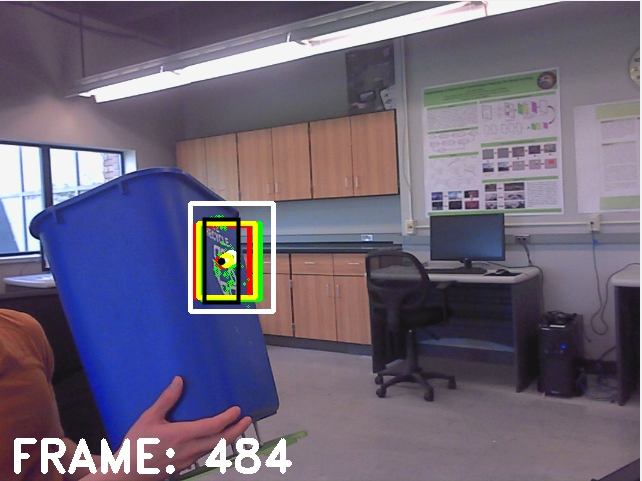}
                \vspace{-1em}
        \end{subfigure}
        \begin{subfigure}[!h]{0.23\textwidth}
                \includegraphics[width=\textwidth]{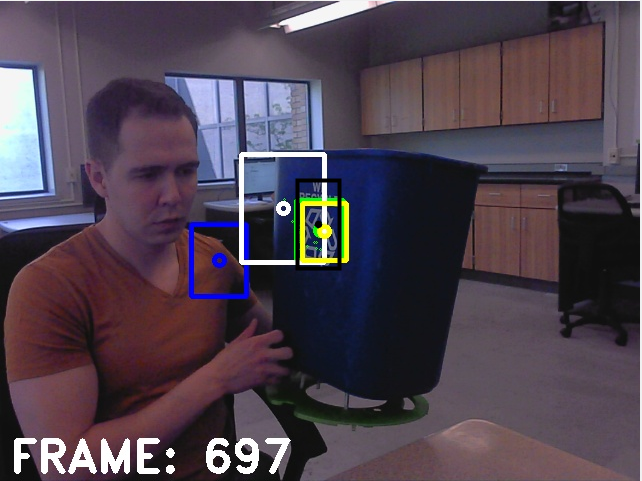}
                \vspace{-1em}
        \end{subfigure}
               \caption{Tracking a recycling bin. Frame 130 shows when DSSTtld loses track due to out-of-plane rotation of the target, while the other approaches continue tracking normally. Frame 301 shows the majority voting taking place. While  Struck and CMT are tracking the recycling bin, DSSTtld could not recover. Frame 484 shows all the approaches working together giving a good estimate before DSSTtld loses track again. At frame 697, Struck drifts and DSSTtld loses track due to an out-of-plane rotation in a previous frame. Despite these problems, HAB-DF is able to keep track of the target for the entire sequence.}\label{fig:track_bin}
\end{figure}

An additional experiment was conducted using a recycling bin as target because of its distinct appearance. Figure~\ref{fig:track_bin} exhibits different images from the experiment. Figure~\ref{fig:trash_bin} shows the different metrics collected during the experiment. Figure~\ref{fig:Jc_ind_11} shows the $J_{IDX}$ for each approach. Up to the $100^{th}$ frame, all approaches have similar performance, with HAB-DF leading in accuracy most of the time. In this scenario, Struck showed better performance, since the object was kept almost at  a constant distance.  It was not until frame $700$ that Struck lost track. Figure~\ref{fig:4D_d_11} shows the Euclidean distance $d$. DSSTtld performed the worst due to pose variations and out-of-plane rotations of the object, while CMT had a reasonable performance throughout the run. Furthermore, the HAB-DF leads in performance among all approaches, relying only on the best detectors at each frame. Figure~\ref{fig:succ_bar_11} shows once again that HAB-DF outperforms  all of the other approaches.  

Figure~\ref{fig:cov_plot_11} displays how the adaptation of the HAB-DF took place. When the distortion of DSSTtld fell below the set threshold, the MD triggered. Between frames 100-300 and 500-800 the detector did not overcome distortions caused by out-of-plane rotations of the object, lowering DSSTtld's confidence, and consequently losing track. CMT showed several spikes caused by substantial delays in processing key points. This behavior does not affect the overall approach, as asynchronous measurements are accounted for by the MD and majority voting.

Figures~\ref{fig:con_spx_11} and \ref{fig:con_spy_11}   illustrate the object position in the frame with respect to the desired set-point ($Sp_u=320$ and  $Sp_v=240$ which are the pixel center coordinates of the image). This graph shows that the experiment was consistent with the motion of the target. Despite some detectors being lost along the experiment, the transition among them was soft.

\subsection{UAV Platform}

\begin{figure}[!h]
        \centering
        \begin{subfigure}[!h]{0.23\textwidth}
                \includegraphics[width=\textwidth]{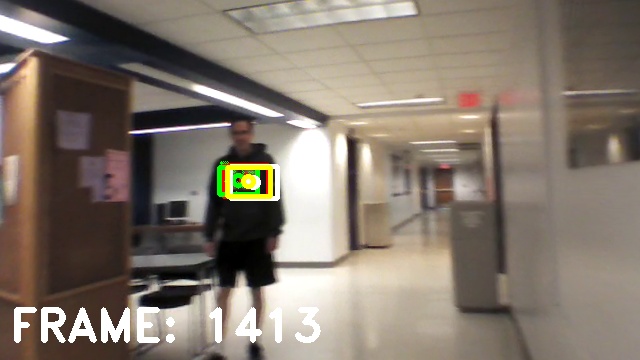}
                \vspace{-1em}
        \end{subfigure}
        \begin{subfigure}[!h]{0.23\textwidth}
                \includegraphics[width=\textwidth]{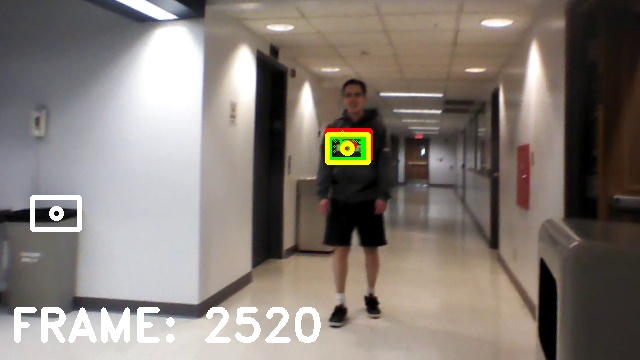}
                \vspace{-1em}
        \end{subfigure}
        ~
        \begin{subfigure}[!h]{0.23\textwidth}
                \includegraphics[width=\textwidth]{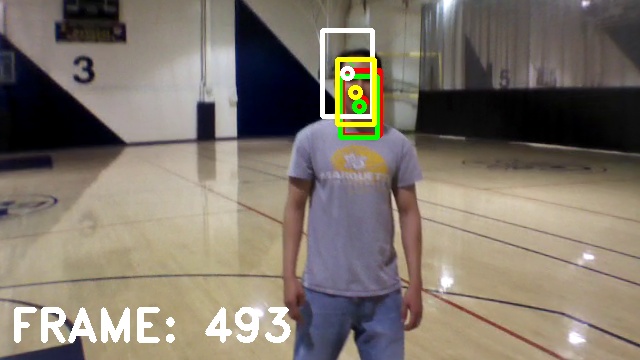}
                \vspace{-1em}
        \end{subfigure}
        \begin{subfigure}[!h]{0.23\textwidth}
                \includegraphics[width=\textwidth]{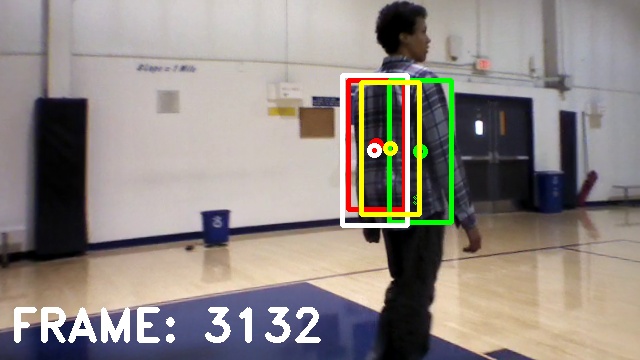}
                \vspace{-1em}
        \end{subfigure}
               \caption{Images from UAV Trials. The frames show all the detectors working properly in the hallway and gym scenarios while the HAB-DF fuses their measurements. During the trial, DSSTtld loses track several times, as illustrated in frame 2520, while CMT and Struck continue to track and HAB-DF properly combines their outputs. Struck shows significant scale disparity, while the combined output correctly estimates the size of the target. Frame 3132 shows a different target in which all three detectors are working albeit with some positional inaccuracy. The combined estimate is more accurate.}\label{fig:Per_tck}
\end{figure}

\begin{figure}[!h]
	\medskip
    \begin{center}
    		\begin{subfigure}[h]{.65\columnwidth}
        	\includegraphics[width=\columnwidth]{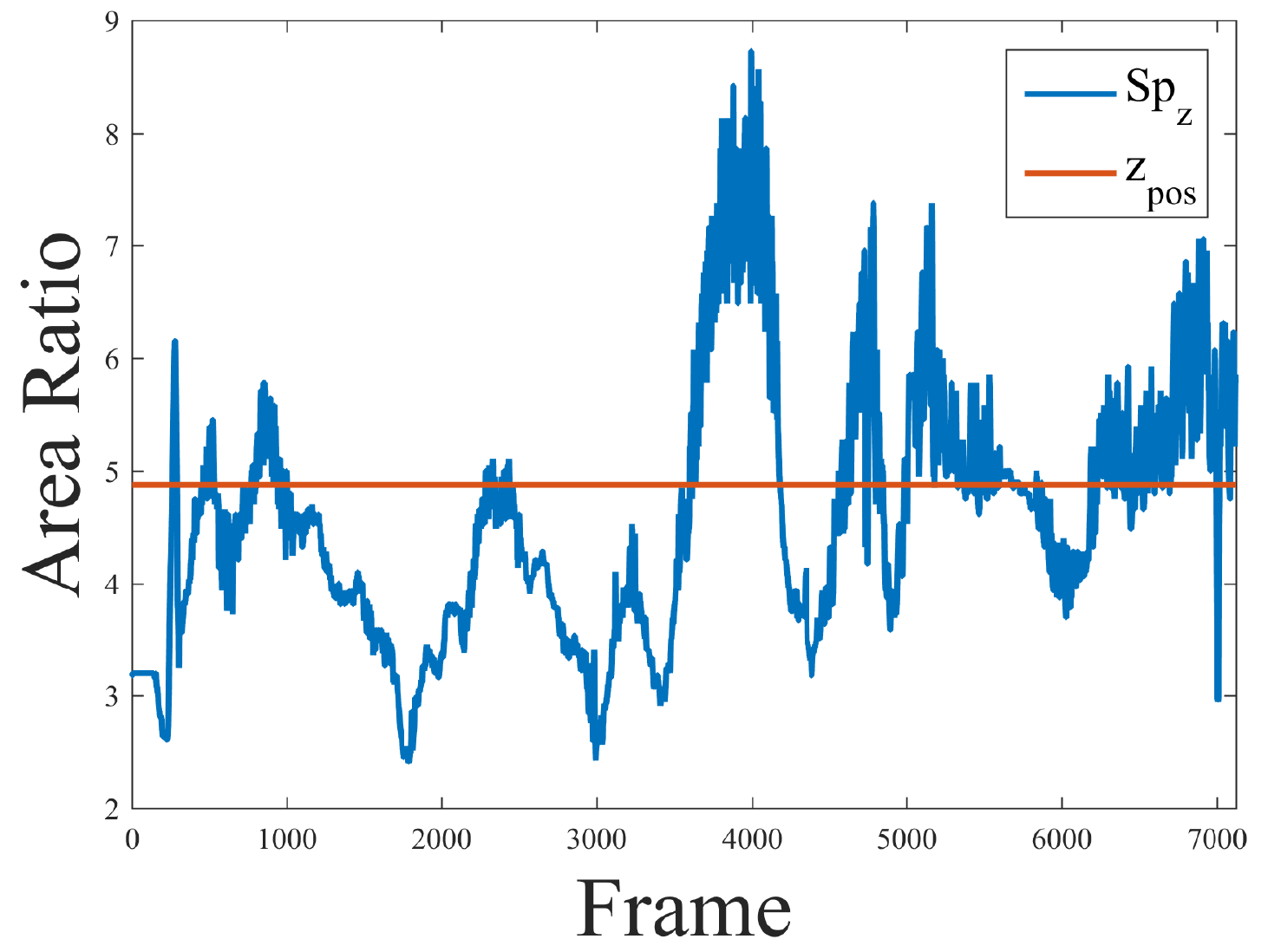}
        \caption{Relative distance to target}
        \label{fig:Depth_D}
    		\end{subfigure}%
    \end{center}
    
    \begin{center}   
    \begin{subfigure}[!h]{.65\columnwidth}     
        \includegraphics[width=\columnwidth]{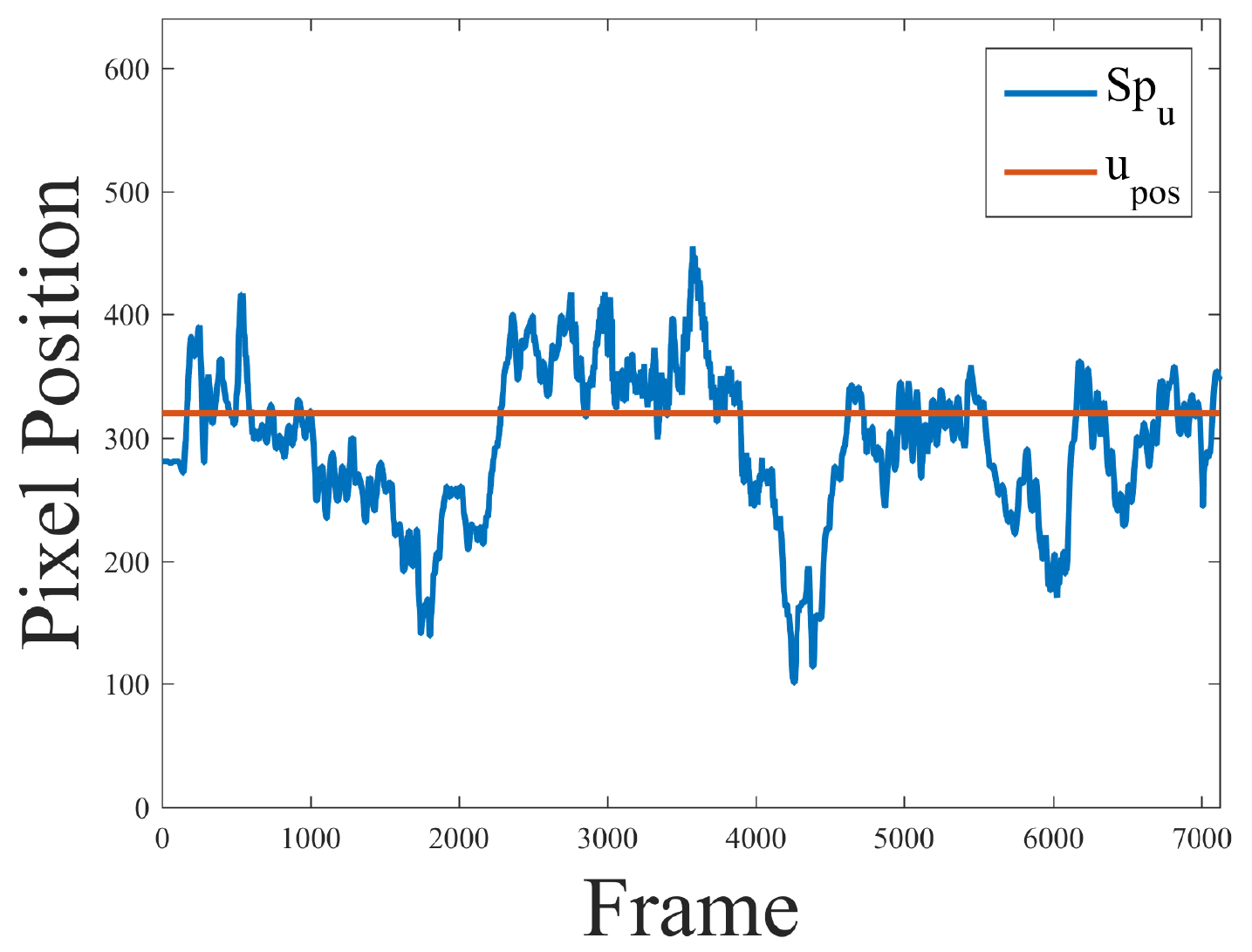}
        \caption{Horizontal target coordinate}  
        \label{fig:Sp_x_D}
    \end{subfigure}
    \end{center}  
    \begin{center}
    \begin{subfigure}[!h]{.65\columnwidth}
        \includegraphics[width=\columnwidth]{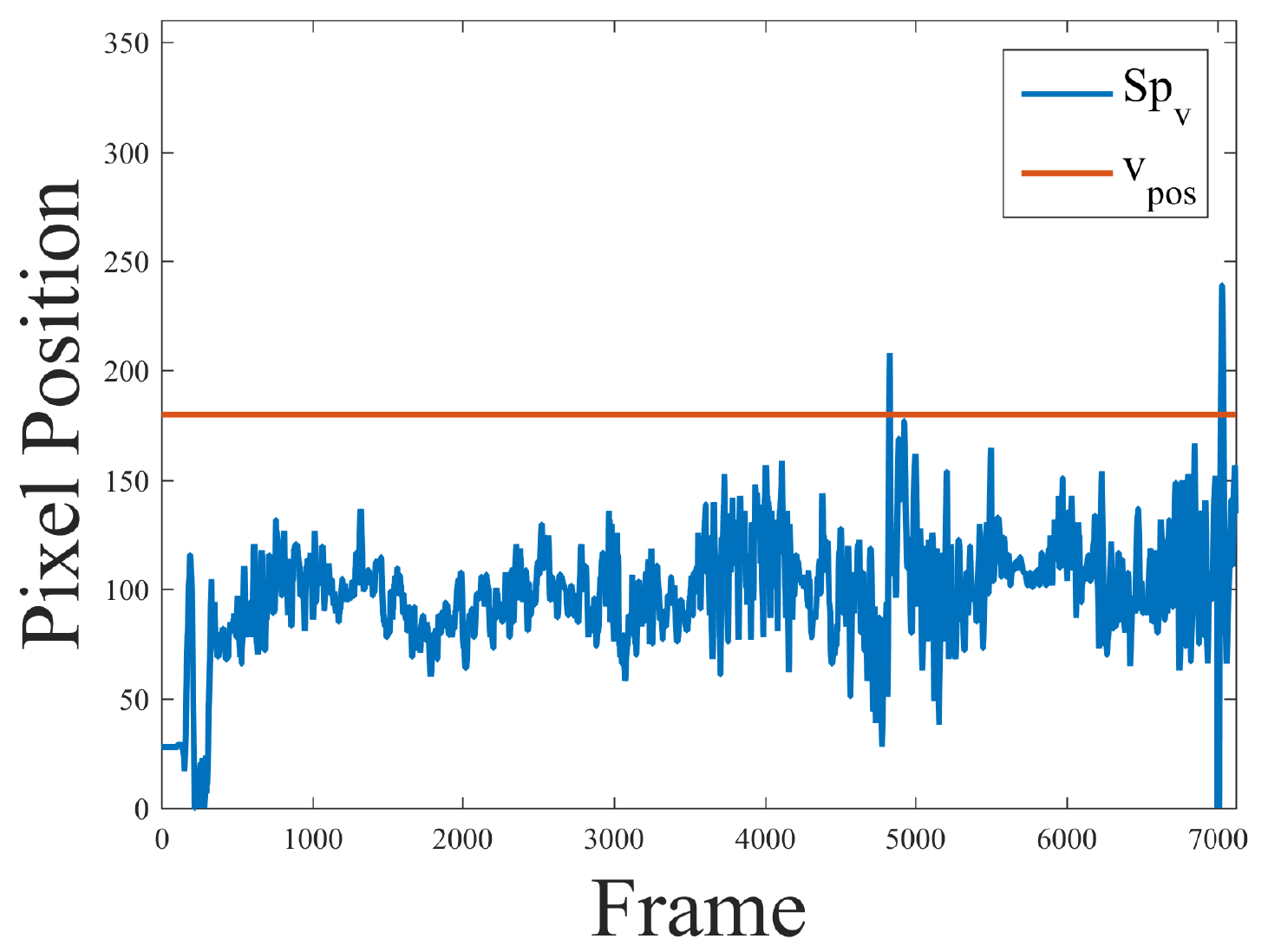}
        \caption{Vertical target coordinate}
        \label{fig:Sp_y_D}
    \end{subfigure}
    \end{center}
    
    \begin{center} 
    \begin{subfigure}[!h]{.65\columnwidth}
        \includegraphics[width=\columnwidth]{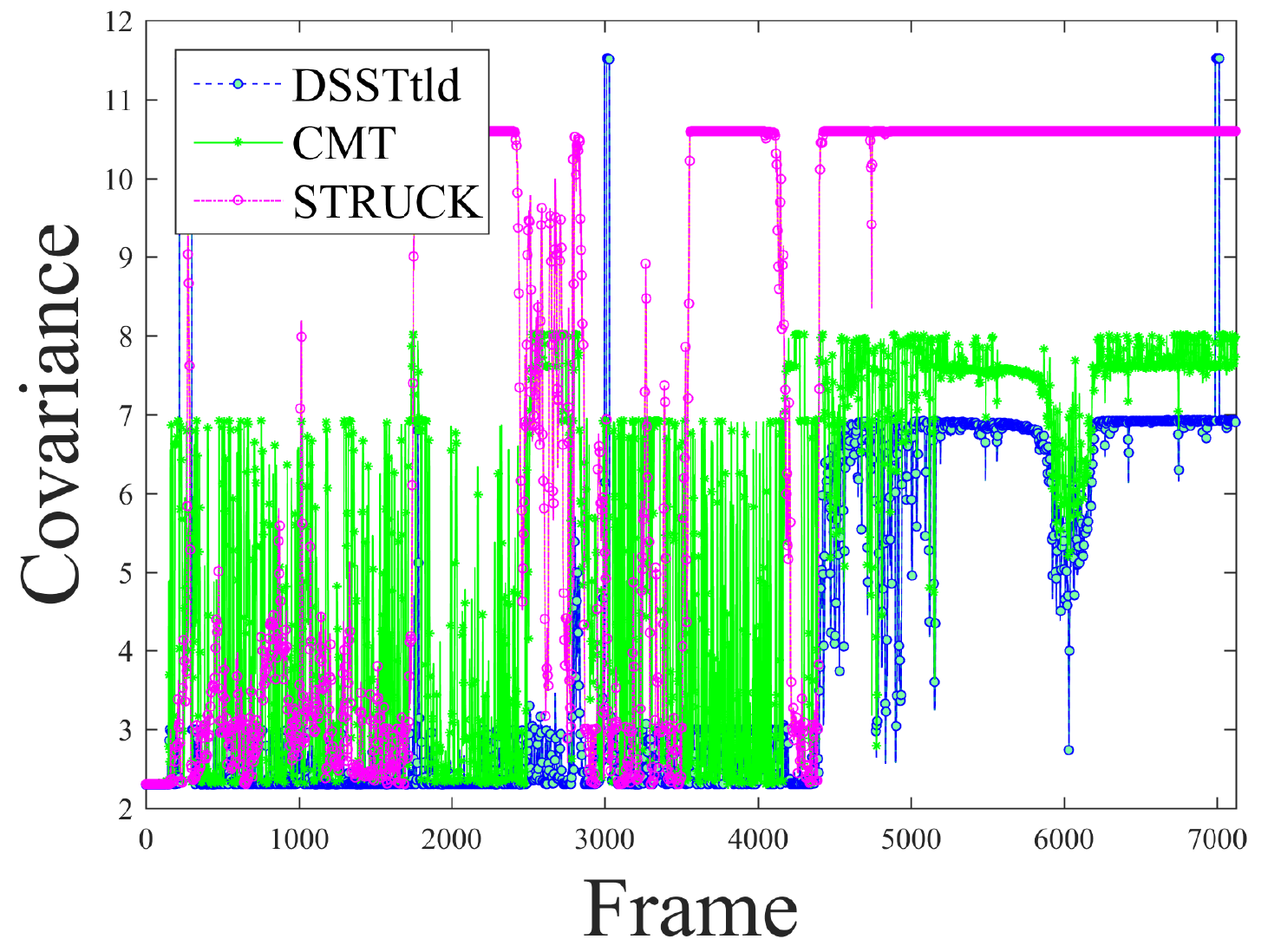}
        \caption{Adaptive covariance}
        \label{fig:Cov_D}
    \end{subfigure} 
    \end{center}
    
\caption{Tracking a person in a gym with a UAV. Figures~\ref{fig:Depth_D}, \ref{fig:Sp_x_D}, and \ref{fig:Sp_y_D} show the behavior of the UAV along the trial. Figure~\ref{fig:Cov_D} shows the adaptive behavior of the HAB-DF during the experiment.}
    \label{fig:Drone_exp}
\end{figure}

Figure~\ref{fig:Per_tck} shows snapshots of experiments using a small UAV. These experiments were carried out indoors and consisted of following several targets in a hallway and in a gym. The results of one of these experiments can be seen in Figure~\ref{fig:Drone_exp}. Figure~\ref{fig:Depth_D} displays the relative distance to the target as estimated by the ratio between the area of the target and the image area. The initial ratio is used as the set point, and the error is used to control the UAV pitch. Figures~\ref{fig:Sp_x_D} and \ref{fig:Sp_y_D} show the vertical and horizontal target positions within the frame and the corresponding set points. The offset observed in Figure~\ref{fig:Sp_y_D}  is due to the coupled effect of the pitch and throttle controllers as the target moves (i.e., as the UAV moves forward, its camera faces down). Although this effect is unavoidable with a fixed camera, it could be resolved with a camera that can be controlled independently from the UAV.  Figure~\ref{fig:Cov_D} shows the amount of penalization suffered by each tracker throughout the trial. It is interesting to note that in this scenario Struck shows improved performance in comparison to the pant-tilt system experiments. This is a result of the fact that the target scale remains approximately constant as the UAV follows it.

Redundant information allows the platform to track the target for longer periods of time. In the sequence shown in Figure~\ref{fig:Drone_exp}, HAB-DF was able to keep track of the target for 7132 frames, until all the detectors lost track of the target simultaneously. In comparison, DSSTtld first lost track at frame 220, Struck at frame 275, and CMT at frame 1738. While these trackers were often able to recover from failure because the target was eventually brought back to the center of the image, had the control actions been taken according to any one of those trackers individually, the platform would likely not have been able to continue following the target. The proposed scheme allows the system to ignore lost detectors and rely on those that provide confident estimates. Failures are evident in Figure~\ref{fig:Cov_D}, which shows to what extent each detector is penalized.

\section{Conclusion}

In this work, a Hierarchical Adaptive Bayesian Data Fusion method was presented. While the algorithm is not limited to specific applications, the main scenario under consideration was vision-based robotic control. The method outperformed single detectors, with better accuracy and kept track for longer periods of time. Moreover, no training data was used while most approaches in this field rely on machine learning techniques, most of which require large amounts of training data for good performance. Even when substantial amounts of training data are available, these methods may be unable to handle situations that were not properly explored during training. The HAB-DF relies instead on the local statistical performance of the individual data sources. In addition, the decentralized architecture allows the experts to operate asynchronously, while penalizing measurements that are delivered to the fusion center with significant delays. Finally, the weighted majority voting scheme allows sensors that provide measurements which are discrepant or have low confidence to be automatically discarded from the estimation.

Moreover, the two platforms tested show that this algorithm is suitable for real-time applications with good performance. Both platforms were able to follow practical objects with different characteristics without any prior training. Additionally, it shows that when detectors/sensors with different performances are combined, they can outperform single methods.

\bibliographystyle{IEEEtran}
\bibliography{IEEEabrv,bibliographypaper}

\end{document}